\documentclass[runningheads]{llncs}
\usepackage[T1]{fontenc}
\usepackage{graphicx}

\usepackage{booktabs}
\usepackage{times}
\usepackage{latexsym}
\usepackage{microtype}
\usepackage{inconsolata}
\usepackage{arydshln}
\usepackage{graphicx}
\usepackage{multirow}
\usepackage{multicol}
\usepackage{booktabs}
\usepackage{amssymb}
\usepackage{bbding}
\usepackage{pifont}
\usepackage{wasysym}
\usepackage{utfsym}
\usepackage{fontawesome}
\usepackage{xcolor}
\usepackage{siunitx}
\usepackage{soul}
\usepackage{pifont}
\usepackage{colortbl} 
\usepackage{makecell}
\usepackage{tcolorbox}

\definecolor{green}{HTML}{ccece7}
\definecolor{green-black}{HTML}{16AB00}
\newcommand{\greentext}[1]{\textcolor{green-black}{#1}}

\definecolor{red}{HTML}{fce3e1}
\definecolor{red1}{HTML}{f4796f}
\newcommand{\redtext}[1]{\textcolor{red1}{#1}}

\definecolor{blue}{HTML}{4857AA}

\begin{document}
\title{Fine-grained User Behavior Simulation on Social Media Based on Role-playing Large Language Models}
%
%
\author{Kun Li\inst{1} \inst{2} \and Chenwei Dai\inst{1} \inst{2} \and Wei Zhou\inst{1} \inst{2} \thanks{* Corresponding author} \and Songlin Hu \inst{1} \inst{2}}
\authorrunning{Kun Li \and Chenwei Dai et al}

\institute{School of Cyber Security, University of Chinese Academy of Sciences, China \and
Institute of Information Engineering, Chinese Academy of Sciences, China \\
\email{\{likun2,daichenwei,zhouwei,husonglin\}@iie.ac.cn}}

\maketitle              
%
\begin{abstract}
Large language models (LLMs) have demonstrated impressive capabilities in role-playing tasks. However, there is limited research on whether LLMs can accurately simulate user behavior in real-world scenarios, such as social media. This requires models to effectively analyze a user's history and simulate their role. In this paper, we introduce \textbf{FineRob}, a novel fine-grained behavior simulation dataset. We collect the complete behavioral history of 1,866 distinct users across three social media platforms. Each behavior is decomposed into three fine-grained elements: object, type, and content, resulting in 78.6k QA records. Based on FineRob, we identify two dominant reasoning patterns in LLMs' behavior simulation processes and propose the \textbf{OM-CoT} fine-tuning method to enhance the capability. Through comprehensive experiments, we conduct an in-depth analysis of key factors of behavior simulation and also demonstrate the effectiveness of OM-CoT approach\footnote{Code and dataset are available at \url{https://github.com/linkseed18612254945/FineRob}}.

\end{abstract}
\section{Introduction}
Large language models (LLMs) have attracted significant attention for their ability to engage in role-playing. These models can be guided by predefined role profiles to generate conversations that align with a character’s speaking style\cite{DBLP:journals/corr/abs-2311-16832}, knowledge\cite{DBLP:conf/acl/Lu0ZZ24}, and personality traits\cite{DBLP:conf/emnlp/ChenWJ0LCWL23}. Recently, numerous agent frameworks have been introduced to extend LLMs' capabilities beyond simple dialogue generation\cite{DBLP:conf/emnlp/WangCC23,DBLP:journals/corr/abs-2402-10618,DBLP:conf/coling/WangDGL24}. However, accurately simulating human-like behaviors poses a substantial challenge, particularly in complex real-world scenarios.

In this paper, we focus on simulating real social media users' behavior. 
Researches have shown that the primary motivations for social network users are self-presentation and self-disclosure\cite{schlosser2020self}. These self-presentations are consciously or unconsciously reflected in all aspects of user behavior\cite{kaplan2010users}.Considering that, we are going to break down each behavior into three fine-grained aspects: \textbf{object} (the target or recipient of the behavior), \textbf{type} (the nature of the behavior), and \textbf{content} (the specific details). An ideal role-playing agent should be able to accurately predict each aspect of user behavior.

\begin{figure}[tp]
    \centering
    \includegraphics[width=\linewidth]{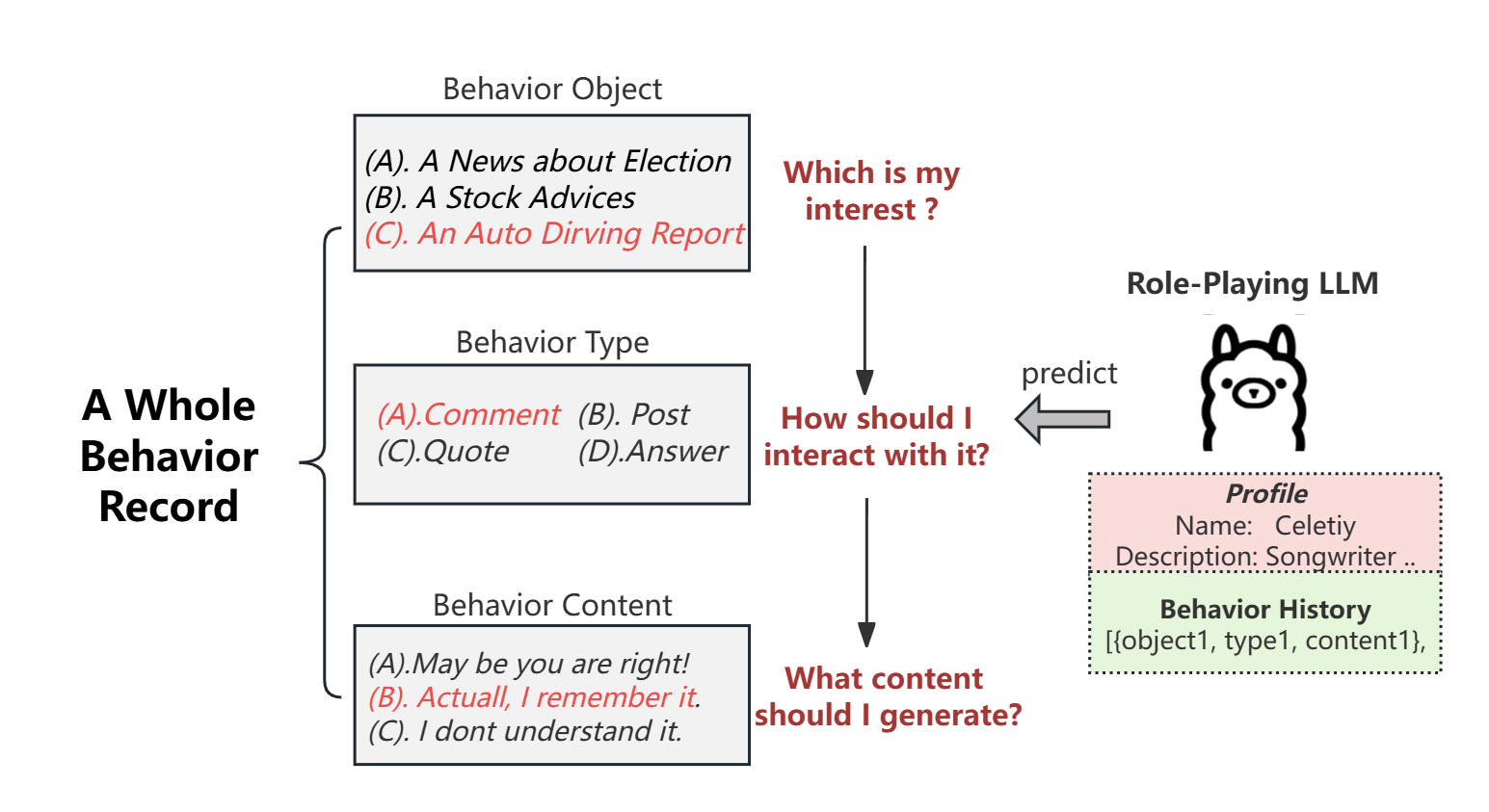}
    \caption{An example of FineRob,  requires LLM to simulat behavior choices that align with a role's profile and historical data. We decompose a complete behavior record into three fine-grained components: selecting the recipient of the action, determining the action type, and specifying the behavior details.}
    \label{fig:example}
\end{figure}

Specifically, we introduce the \textbf{FineRob} (\textbf{Fine}-Grained \textbf{Ro}le \textbf{B}ehavior) dataset, a novel benchmark for role-behavior simulation on social media. FineRob collects real user behavior data from three major platforms—Twitter, Reddit (primarily English), and Zhihu (primarily Chinese)—encompassing 1,866 distinct users and 78.6k fine-grained behavior QAs. Each raw user behavior is decomposed into three aspects. For example, when a user comments on a post, he first chooses a target (e.g., a post or article from their feed), then decides on the type of action (e.g., comment, like, or share), and finally generates content that aligns with his persona, as illustrated in Figure \ref{fig:example}.

Using the FineRob dataset, we conduct a comprehensive evaluation of nine widely used LLMs. We analyze the reasoning processes of LLMs and find two prevalent reasoning patterns: "\textbf{role stereotype-based reasoning}" and  "\textbf{observation and memory-based reasoning}". The former tends to over-rely on user profile information, which leads to reduced accuracy in behavior simulation. In contrast, the latter compares current observations with past behaviors, resulting in more accurate simulations, which is favored by more advanced models, such as GPT-4o. Building on this insight, we introduce a novel fine-tuning approach, OM-CoT, which utilizes special tokens to explicitly incorporate observation and memory analysis into the reasoning process. 
 Extended experiments confirm the effectiveness of our approach.

The contributions of this study can be summarized as follows:

\begin{itemize}
    \item \textbf{We introduce the FineRob dataset}, consisting of 78.6k fine-grained behavior element simulation QA records from 1866 real-world social media users. FineRob serves as a realistic, multilingual benchmark for evaluating LLMs’ ability to simulate role-specific behaviors.
    
    \item \textbf{We comprehensively assess behavior simulation} across nine mainstream LLMs, focusing on the reasoning patterns employed during simulation.
    
    \item \textbf{We propose the OM-CoT, a novel fine-tuning method} that explicitly integrates observation and memory analysis into the reasoning process using special tokens. Our experiments demonstrate significant improvements across all three behavior element simulation tasks.
\end{itemize}

\begin{table}[h]
    \caption{Comparison between FineRob and previous role-playing datasets.}
    \label{table:datasets}
    \centering
    \resizebox{\textwidth}{!}{
    \begin{tabular}{l|l|l|l|l|l|c|c}
        \toprule
        \textbf{Dataset} & \textbf{Source} & \textbf{Size} & \textbf{Usage} & \textbf{Language} & \textbf{Type} & \textbf{isReal?} & \textbf{Fine-Grained} \\
        \midrule
        PersonalDialog\cite{DBLP:journals/corr/abs-1901-09672} & Weibo & 20.8M & Train\&Test & ZH & Dialogue & $\checkmark$ & \textbf{--} \\
        Ditto\cite{DBLP:conf/acl/Lu0ZZ24} & LLM Synthetic & 4k & Train\&Test & EN,ZH & Dialogue & $\times$ & \textbf{--} \\
        LaMP-7\cite{DBLP:conf/acl/SalemiMBZ24} & Twitter & 12k & Train\&Test & EN & Dialogue & $\checkmark$ & \textbf{--} \\
        PIPPA\cite{DBLP:journals/corr/abs-2308-05884} & Character.AI & 26k & Train & EN & Dialogue & $\times$ & \textbf{--} \\
        \cdashline{1-8}
        ROCStories\cite{DBLP:conf/naacl/MostafazadehCHP16} & Little Stories & 98K & Train\&Test & EN & Behavior & $\times$ & $\times$ \\
        choices13K\cite{DBLP:conf/icml/PetersonB0GR19} & gamble & 13K & Test & EN & Behavior & $\checkmark$ & $\times$ \\
        Life-Choice\cite{DBLP:journals/corr/abs-2404-12138} & Novels & 1.4K & Test & EN & Behavior & $\times$ & $\times$ \\
        \textbf{FineRob} & SocialMedias & 78.6K & Train\&Test & EN,ZH & Behavior & $\checkmark$ & $\checkmark$ \\
        \bottomrule
    \end{tabular}
    }

\end{table}

\section{Related Work}

\subsection{Role-Playing LLM}
Recently, numerous RP-LLMs have been designed for conversational applications and have already found commercial uses like Character.ai\footnote{\url{https://character.ai/}}. Researchers collect a wide range of dialogue datasets to support the study, including the data from real-life individuals\cite{DBLP:conf/acl/GaoLZFW23,DBLP:conf/iclr/DinanRSFAW19} or fictional characters from novels \cite{DBLP:conf/emnlp/ChenWJ0LCWL23,DBLP:conf/acl/AhnLLKYLK24}. Beyond this, techniques such as in-context learning (ICL)\cite{DBLP:conf/eacl/ZhaoZLZLHG24} and retrieval-augmented generation (RAG)\cite{DBLP:conf/acl/LiuCFMM23} have been employed. Additionally, supervised fine-tuning on targeted dialogue datasets\cite{DBLP:conf/emnlp/ShaoLDQ23} and methods like LoRA\cite{DBLP:journals/corr/abs-2402-13717} have further enhanced RP-LLMs' role-playing capabilities. Despite these advancements, research in RP-LLMs remains in its early stages, with a primary focus on mimicking conversations.


\begin{figure*}[ht]
    \centering
    \includegraphics[width=\linewidth]{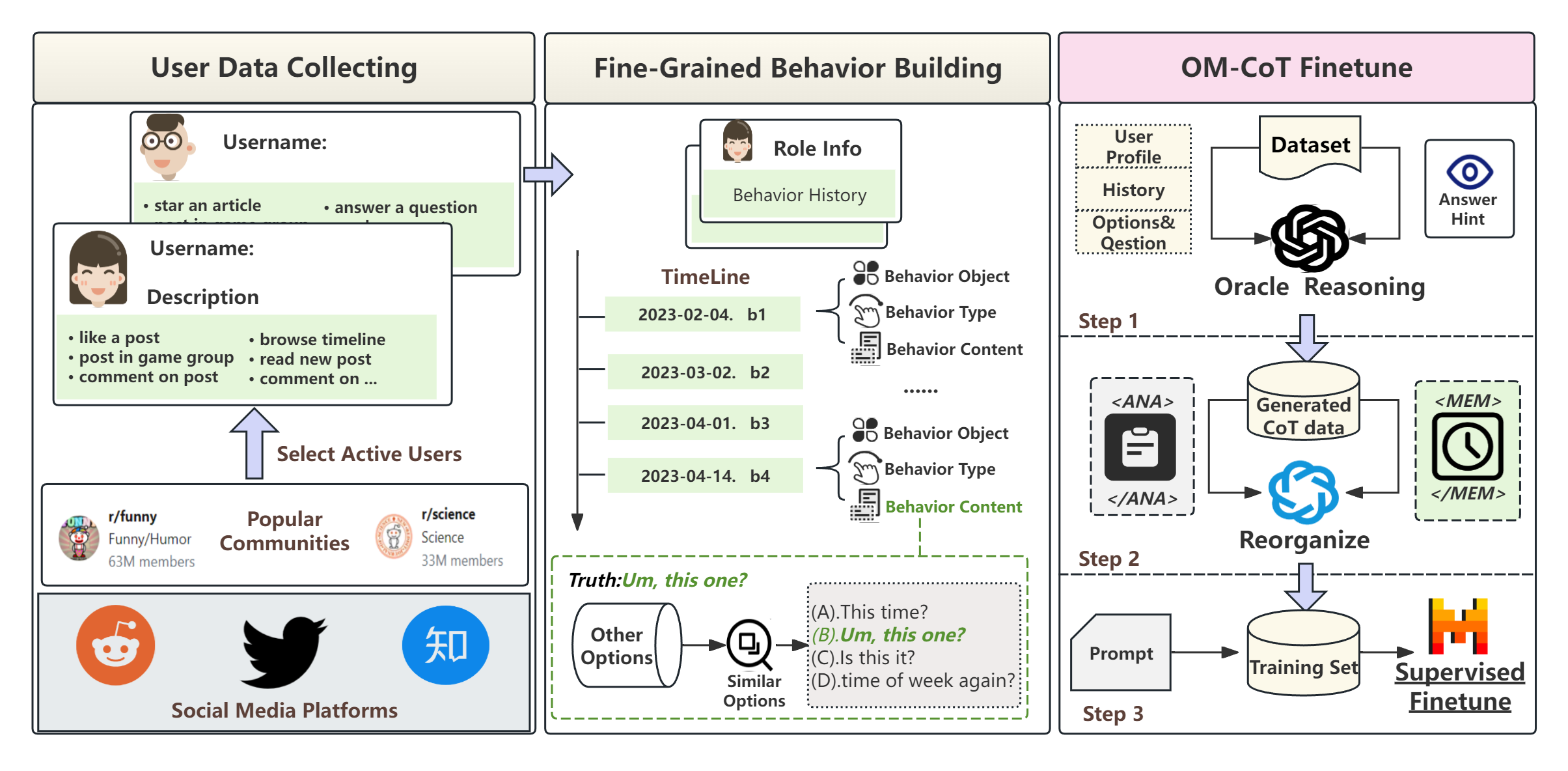}
    \caption{Overview of our work, The left and middle sections of the figure illustrate the process of constructing the FineRob dataset. The right section shows how OM-COT-FineTune training details, including data augmentation, reorganize with special tokens and SFT training.}
    \label{fig:framework}
\end{figure*}

\subsection{LLM for User Behavior simulation}
Researchers have recognized that LLMs are not only adept at mimicking conversation but also capable of simulating complex behaviors\cite{DBLP:conf/uist/ParkOCMLB23,DBLP:conf/emnlp/WangCC23,DBLP:journals/corr/abs-2406-11683,DBLP:conf/coling/WangDGL24}. For instance, \cite{wang2023user} demonstrated that LLMs can mimic real users' preferences to movie recommendations. \cite{DBLP:journals/corr/abs-2405-13362} explored the use of reinforcement learning algorithms to optimize recommendation systems based on user feedback simulated by LLMs. The work by \cite{DBLP:journals/corr/abs-2404-12138} is particularly relevant, which introduced the "LIFECHOICE" dataset to assess LLMs' ability to make broad, macro-level decisions in fictional contexts. In contrast, our research focus on the LLMs' capacity to simulate fine-grained, micro-level behaviors in real world scenario, offering a more detailed analysis.


\section{FineRob Dataset}


\subsection{Data Collecting}



Our goal is to explore how LLMs simulate the behaviors of real internet users. To achieve this, we focused on social media platforms, including Twitter (now X), Reddit, and the Chinese question-and-answer website Zhihu. From these platforms, we collected extensive behavioral histories of real users, as shown in the first part of Figure \ref{fig:framework}. Unlike other role-playing tasks, we emphasize fine-grained behavior simulation in real-world scenarios. A detailed comparison is provided in Table\ref{table:datasets}.

\paragraph{Principles} Our data collection strategy is guided by several key principles. (1) \textbf{Popularity}: We focused on mainstream, widely discussed topics and scenarios to ensure the dataset reflects a representative sample of user behaviors. (2) \textbf{Diversity}: We include a broad range of user profiles and behavioral patterns to enhance the generalizability of our findings.(3)\textbf{Activity}: We select users who are active within the community and engage in various types of behavior, helping to minimize data contamination from social bots, fake accounts, or other non-human users.

\paragraph{User Selection} 
To collect our dataset, we target active users by selecting them from trending topics or communities displayed on aggregation page of each platform. For Reddit, we choose popular posts from the top 20 communities and filter participants based on predefined principles. 
 Using the PRAW\footnote{\url{https://praw.readthedocs.io/en/stable/}},  we automatically collect these users’ timelines, including their complete historical behaviors. For Twitter, we purchase access to the official API\footnote{\url{https://developer.x.com/en/docs/x-api}} and prioritize users who actively engaged in discussions on trending topics.
On Zhihu, we focus on users who frequently ask or answer questions. To ensure reliability, we filter users with at least 70 accessible historical behaviors. Conversely, we exclude users who exhibit excessive activity, as these accounts may be operated by multiple individuals. An user example with behavior history can be found in Table \ref{appendix: record-case}.

\begin{table}[htp]
\caption{Valid behavior types for LLM agents on social medias.}
\label{appendix table: behavior types}
\centering
\resizebox{\textwidth}{!}{
\begin{tabular}{cccc}
\toprule
Name & Description & \multicolumn{1}{l}{Need Target?} & \multicolumn{1}{l}{Need Content?} \\
\midrule
\rowcolor{gray!20} \multicolumn{4}{c}{Reddit} \\
comment & comment to post or other comment on reddit & $\checkmark$ & $\checkmark$ \\
post & create a new post on subreddit & $\times$ & $\checkmark$ \\
\rowcolor{gray!20} \multicolumn{4}{c}{Twitter} \\
replied to & replied to other tweets or comments & $\checkmark$ & $\checkmark$ \\
post & create a new tweet & $\times$ & $\checkmark$ \\
like & approval or support for a tweet without sharing. & $\checkmark$ & $\times$ \\
quoted & Adds your comment to someone else's shared tweet. & $\checkmark$ & $\checkmark$ \\
retweet & shares someone else's tweet with your followers unchanged. & $\checkmark$ & $\times$ \\
\rowcolor{gray!20} \multicolumn{4}{c}{Zhihu} \\
new question & Ask a question to seek answers & $\checkmark$ & $\times$ \\
answer & Answer a question to share knowledge & $\checkmark$ & $\checkmark$ \\
opinion & Post a thought to share opinions & $\times$ & $\checkmark$ \\
post article & Post a new article & $\times$ & $\checkmark$ \\
update question & Update a question & $\checkmark$ & $\checkmark$ \\
agree answer & Agreed with an answer by upvoting to show support or approval. & $\checkmark$ & $\times$ \\
follow question & Follow a question to receive updates or answers about it. & $\checkmark$ & $\times$ \\
agree article & Agreed with an article by upvoting to show support or approval. & $\checkmark$ & $\times$ \\
bookmark article & Saved an article to bookmark it for later reference or reading & $\checkmark$ & $\times$ \\
bookmark answer & Saved an answer to bookmark it for later reference or reading & $\checkmark$ & $\times$ \\
approve answer & Approve an answer by endorsing its accuracy or helpfulness & $\checkmark$ & $\times$ \\
\bottomrule
\end{tabular}
}

\end{table}


\subsection{Fine-Grained Behavior Building}
Next, we convert the raw user timelines into a fine-grained behavior simulation QA dataset with multiple-choice format. Specifically, each behavior record is broken down into three elements: \textbf{object} (the recipient of the behavior), \textbf{type} (the nature of the behavior), and \textbf{content} (the specific details of the behavior). This process is illustrated in the middle section of Figure 2. \ref{fig:framework}.

A significant challenge lies in constructing valid alternative options for each behavior element in the multiple-choice format. The behavior \textbf{type} options are relatively straightforward, as platforms typically have predefined actions like "Post" and "Comment." These behavior types can be found in Table \ref{appendix table: behavior types}. For the behavior's object and content options, we construct a candidate set based on the user's active times and communities. From this set, we calculate the similarity to the correct option using sentence embeddings\footnote{\url{https://huggingface.co/moka-ai/m3e-base}}. To increase task difficulty and introduce ambiguity, we randomly sample three options with closely aligned sentiment to the correct answer.

Finally, we collected a total of 78.6k behavior element simulation records and split the dataset into approximately 61k for the training set and 17.6k for the test set. To assess the generalization capability of large language models (LLMs), we ensured that no user roles overlap between the training and test sets.



\begin{figure}[h]
    \centering
    \includegraphics[width=\linewidth]{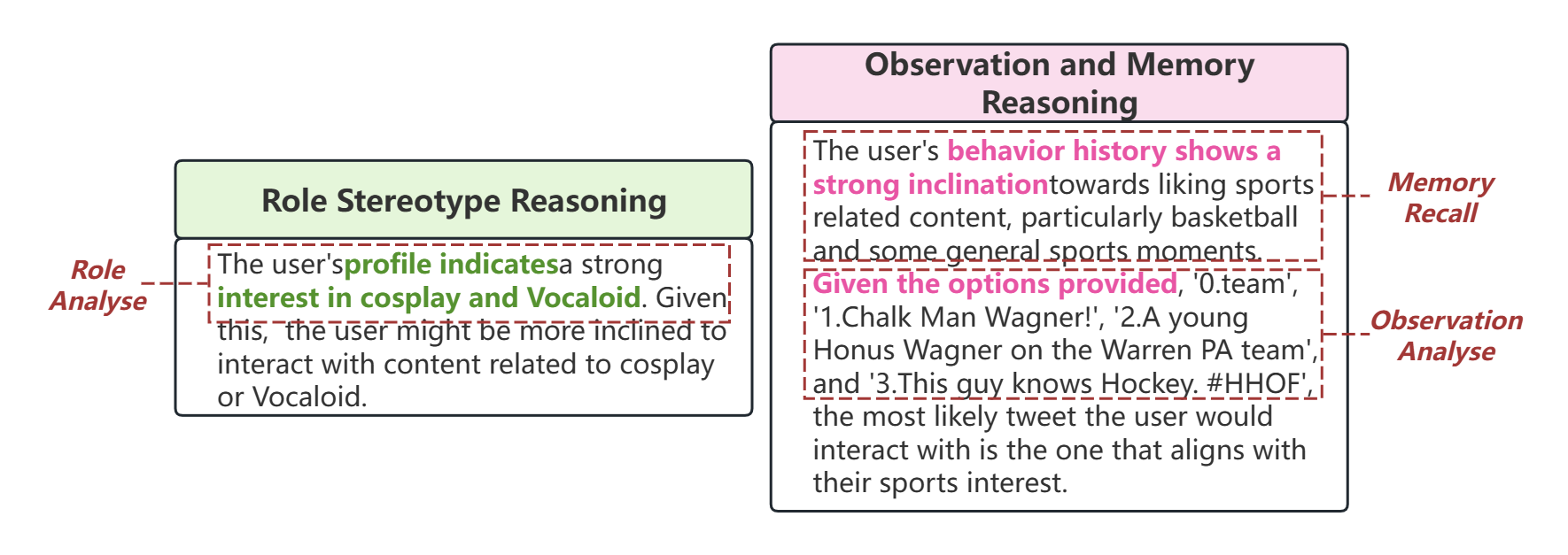}
    \caption{Two typical patterns of COT reasoning for behavior simulation. The "Role Stereotype" pattern focus on role analysis. The "Observation and Memory" pattern simulats future behavior by considering the relationship between the character's history and observed options.}
    \label{fig:pattern}
\end{figure}


\section{Methodology}

\begin{figure}[]
    \centering
    \includegraphics[width=\linewidth]{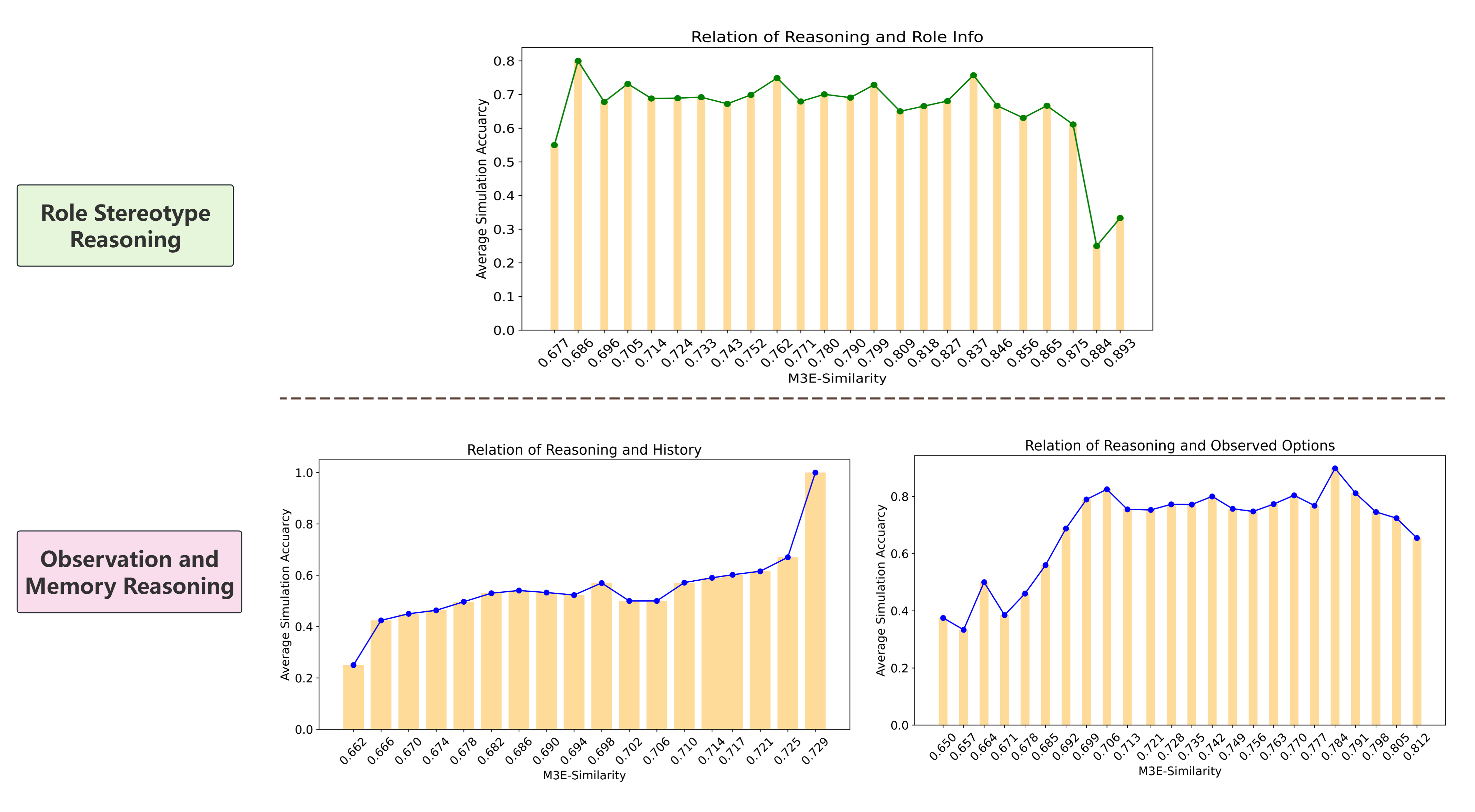}
    \caption{Analysis of simulation accuracy changes across different similarity levels between reasoning and various parts of the prompt. The results are generated using ChatGPT-3.5-turbo-0125 on the Twitter test set, with the average F1-score calculated across three behavior element tasks.}
    \label{fig:compare}
\end{figure}

\subsection{Preliminary Analysis}
We conduct preliminary experiments using a zero-shot Chain of Thought (CoT) approach. Our goal is to understand the reasoning processes LLMs use in behavior simulation tasks. Our analysis reveal two primary reasoning patterns. The first, termed role stereotype-based reasoning, derives outcomes by  analyzing character profiles. The second, \textbf{observation and memory-based reasoning}, involves analysing all observed options and linking them to similar past scenarios. Examples of these patterns are illustrated in Figure \ref{fig:pattern}.

In our comparison, we find that more advanced models, such as GPT-4o, tend to prefer the "observation and memory-based reasoning" pattern, leading to more accurate simulations. 
 To investigate this further, we conduct a quantitative analysis of the similarity between CoT reasoning text and each parts of prompt: behavior history, observed options, and role info. The results, shown in Figure \ref{fig:compare}, reveal an interesting insight:  A higher similarity to role profiles, which may involve more character analysis, does not always lead to better behavior simulations. Instead, focusing on historical data and observed options proves to be more effective in improving simulation accuracy.

\subsection{OM-CoT Finetune}
Based on these findings, we propose a straightforward yet effective method called OM-CoT Finetune (Observation and Memory-based Chain of Thought Finetune) to enhance behavior simulation accuracy in LLMs. This method promotes the "observation and memory-based reasoning" pattern by explicitly integrating observation and memory analysis into the CoT reasoning. The approach involves three steps, as illustrated in the right part of Figure \ref{fig:framework}.

\paragraph{Oracle CoT Generation} 
First, we use a powerful large language model to generate CoT reasoning. To prevent error propagation, we adopt an oracle setting, where the correct answer is provided in the input prompt. This ensures the model references the correct behavior during reasoning. We carefully adjust the prompt to make sure the generated CoT does not inadvertently reveal the correct answer.

\paragraph{Reorganize CoT with special tokens} Next, we introduce two special tokens: \textbf{\texttt{<ANA>}} and \textbf{\texttt{<MEM>}}. A smaller LLM reorganizes the CoT results by wrapping observation-based analysis within \textbf{\texttt{<ANA>}}\textbf{\texttt{</ANA>}} and historical behavior analysis within \textbf{\texttt{<MEM>}}\textbf{\texttt{</MEM>}}. At the end of each reasoning process, the model explicitly states the final behavior decision (e.g., "Therefore, the behavior type is A.Comment").

\paragraph{SFT with Enhanced Dataset} 
Finally, we perform Supervised Fine-Tuning (SFT) on the LLM using the reorganized dataset. The training optimizes for standard language model loss, guiding the model to effectively utilize the special tokens along with system prompts to control the CoT process. We create 60K instruction training data for OM-CoT fine-tuning, which is also available in the codebase.

\section{Experiment}

\subsection{Settings}

\paragraph{Models}
We evaluate a total of nine large language models on FineRob. This includes three commercial LLMs,  ChatGPT-3.5-turbo-0125/GPT-4o-mini/GPT-4o\footnote{\url{https://platform.openai.com/docs/models}},as well as six open-source LLMs:: Mistral-7b-Instruct\footnote{\url{https://huggingface.co/mistralai/Mistral-7B-Instruct-v0.2}}\cite{DBLP:journals/corr/abs-2310-06825},Llama3-8b-Instruct\footnote{\url{https://huggingface.co/meta-llama/Meta-Llama-3-8B-Instruct}}\cite{DBLP:journals/corr/abs-2407-21783}, Solar-10.7b-Instruct\cite{DBLP:journals/corr/abs-2312-15166}, Yi-1.5-34B-Chat\cite{young2024yi}, 
Baichuan2-13B-Chat\footnote{\url{https://huggingface.co/baichuan-inc/Baichuan2-13B-Chat}}\cite{DBLP:journals/corr/abs-2309-10305},and Qwen2-72B-instruct\footnote{\url{https://huggingface.co/Qwen/Qwen2-72B-Instruct}}. The last two models is specialize for Chinese context.

\paragraph{Baselines} We conduct extended experiments on the Mistral-7b-Instruct and Solar-10.7b-Instruct models using four baseline methods: zero-shot, few-shot, standard-CoT fine-tune, and OM-CoT fine-tune. In the few-shot setup, we include a reasoning example created by GPT-4o, which follows the "observation and memory-based reasoning" pattern. The standard-CoT fine-tuning method uses un-reorganized CoT data without special tokens. By comparing these approaches, we aim to evaluate how different training and prompting methods affect LLM behavior simulation performance.

\paragraph{Prompts}
The prompts were similarly structured across all baseline methods and consisted of four main parts: (1) a task description instructing the model to simulat three behavior elements while role-playing a specific character, (2) the role’s profile, which included username, self-description, and areas of interest, (3) behavior history, detailing the target, type, content, and timing of past behaviors, and (4) method-specific instructions and output format requirements. For example, in OM-CoT, the model was instructed to use the \textbf{\texttt{<ANA>}} and \textbf{\texttt{<MEM>}} tokens for analysis based on observation and memory.

\paragraph{Implementation Details}
We utilize LoRA \cite{DBLP:conf/iclr/HuSWALWWC22} for efficient parameter fine-tuning, setting $\alpha$ to 1.0 and $\beta$ to 0.025. All training are conducted with fp16 mixed-precision on 4 $\times$ A100 GPUs over 10 epochs with LLama-factory\footnote{\url{https://github.com/hiyouga/LLaMA-Factory}}\cite{zheng2024llamafactory}. For inference, we use vLLM\footnote{\url{https://github.com/vllm-project/vllm}} to accelerate the process, employing sampling decoding with a temperature of 0.1. The F1 score serves as the evaluation metric across all experiments. To mitigate the impact of randomness inherent in LLMs, we run three trials and compute the mean and standard deviation, ensuring more reliable results.

\begin{table}
    \caption{F1-scores of nine LLMs under a zero-shot setting, including Behavior Object, Behavior Content, and Behavior Type. The average and standard deviation were recorded over multiple runs. The best and second-best results from Commercial-LLM and Open-LLM are highlighted using \textbf{bold} and \underline{underline} formatting.}
    \label{table:main_zeroshot}
    \centering
    \resizebox{\textwidth}{!}{ 
    \begin{tabular}{l|ccc|ccc|ccc}
        \toprule
        & \multicolumn{3}{c|}{\textbf{Reddit}} & \multicolumn{3}{c|}{\textbf{Twitter}} & \multicolumn{3}{c}{\textbf{Zhihu}} \\
        & Object & Content & Type & Object & Content & Type & Object & Content & Type \\
        \midrule
        Random & 10.32 & 10.46 & 6.31 & 10.45 & 8.65 & 11.85 & 10.14 & 10.22 & 3.12 \\
        \rowcolor{gray!20} \multicolumn{10}{c}{\textbf{Commercial-LLM}} \\
        chatgpt-3.5-turbo-0125 & 19.99±0.2 & 19.72±0.1 & \underline{51.22±0.0} & \underline{54.15±0.2} & 37.62±0.0 & \underline{62.33±0.4} & 25.09±0.0 & \underline{33.17±0.0} & 19.24±0.0 \\
        GPT-4-mini & \underline{26.49±0.0} & \underline{23.91±0.0} & 48.14±0.0 & \textbf{63.33}±0.0 & \underline{41.26±0.0} & 85.99±0.0 & 31.65±0.0 & 31.94±0.0 & \underline{24.04±0.0} \\
        GPT-4o & \textbf{28.34±1.2} & \textbf{24.13±0.0} & \textbf{58.14±0.0} & 53.70±0.0 & \textbf{52.90±0.0} & \textbf{86.97±0.0} & \textbf{36.34±0.1} & \textbf{41.49±0.0} & \textbf{26.35±0.0} \\
        \rowcolor{gray!20} \multicolumn{10}{c}{\textbf{Open-LLM}} \\
        Mistral-7b-v2.0 & 25.95±2.0 & 19.62±1.4 & \textbf{22.92±0.7} & 19.99±2.4 & 27.04±6.5 & 62.92±0.9 & 21.56±1.2 & 19.53±1.6 & 10.38±0.2 \\
        LLama3-8b & 10.80±0.5 & 11.8±0.5 & 13.6±0.6 & 22.72±0.9 & 16.23±3.5 & 52.90±0.7 & 21.70±0.9 & 12.66±1.2 & 10.75±0.2 \\
        Solar-10.7b & \textbf{27.31±1.6} & \textbf{25.31±2.3} & 20.52±0.8 & \textbf{53.40±1.6} & 27.72±7.6 & 71.12±1.4 & 19.48±1.6 & 15.00±2.1 & 10.66±0.5 \\
        Baichuan-13b & \underline{27.13±1.4} & 21.98±1.1 & \underline{21.16±0.8} & 16.22±0.6 & 12.64±0.3 & 37.51±1.1 & \textbf{31.27±0.8} & \underline{24.62±1.5} & 15.97±0.5 \\
        yi-34b & 25.94±1.9 & \underline{24.94±1.8} & 18.26±1.6 & 51.14±0.7 & \underline{32.74±2.9} & \textbf{73.98±0.8} & \underline{33.93±1.1} & \textbf{26.84±1.2} & \underline{16.15±0.4} \\
        qwen-2-72b & 12.37±0.7 & 11.19±0.7 & 15.68±2.5 & \underline{52.44±6.7} & \textbf{45.39±5.5} & \textbf{77.65±6.1} & 30.92±0.6 & 22.35±1.4 & \textbf{16.80±0.3} \\
        \bottomrule
    \end{tabular}
    }
\end{table}

\subsection{Main Result}
We first compare the behavior simulation capabilities of the main LLMs under the same zero-shot setting, as shown in Table \ref{table:main_zeroshot}.  Following this, we perform extended experiments to highlight the advantages of the OM-CoT fine-tuning method, with the results presented in Table \ref{table:baselines}. Next, we will discuss some conclusions drawn from the main results.

\paragraph{Commercial Closed-Source Models Still Perform Better.}
While many open-source models have recently demonstrated strong performance on general leaderboards\footnote{\url{https://huggingface.co/spaces/open-llm-leaderboard}}, a detailed comparison in Table \ref{table:main_zeroshot} shows that even the best open-source LLMs lag behind the GPT-4 series by approximately 15\% in average F1 score. Comparable performance is only observed on the Reddit dataset for Behavior Object and Behavior Content simulation tasks. This suggests that role-playing and behavior simulation may require more than just the reasoning abilities. These tasks likely involve advanced capabilities such as empathy and reflective analysis of past behaviors.

\paragraph{Bigger Models Are Not Always Better.}
We observe that larger open-source models do not consistently  outperform smaller alternatives, which is unexpected. For example, Qwen-2-72B, one of the top open-source models, shows strong performance on multilingual tasks, especially with the Zhihu and Twitter datasets. 
 However, its results on the Reddit dataset fell significantly below expectations. In contrast, smaller models like Solar-10.7B and Mistral-7B, despite having fewer parameters, deliver more balanced and competitive outcomes across a broader range of tasks.

\paragraph{OM-CoT Fine-Tuning Enhances Behavior simulation Performance.}
As shown in Table \ref{table:baselines}, we apply OM-CoT fine-tuning to two models with different parameter sizes, a reasoning case can be found in Table \ref{appendix: reasoning-case}. For the Mistral-7B model, performance improve across all nine sub-tasks, with an average F1 score increase of approximately 4.5\%. Similarly, the Solar-10.7B model exhibited gains in seven out of nine tasks, including a significant 9.8\% improvement in the Reddit Behavior Object simulation task. Furthermore, baseline analysis reveal that incorporating examples from the fine-tuning dataset into the few-shot setting do not produce the expected improvements; in some cases, it even perform worse than the zero-shot setting. On the other hand, both Std-CoT-FT and OM-CoT-FT consistently outperform models that were not fine-tuned. These results indicate that, even with different users in the training and test sets, large models can effectively learn generalizable reasoning patterns.

\paragraph{LLMs Struggle with Short-Behavior Tasks, Even When Fine-Tuned.}
A notable result emerge from the Reddit dataset, particularly in the behavior content simulation task, where fine-tuning methods failed to yield performance improvements. Upon further investigation, we find that Reddit content is often brief and lacks clear indicators of user characteristics. This suggests that current language models still struggle to differentiate subtle variations in tone and punctuation within behaviors (e.g., "Good work" vs. "Pretty Nice!!").


\begin{table}
    \caption{F1-scores of OM-COT-FT and other baselines}
    \label{table:baselines}
    \resizebox{\textwidth}{!}{ 
    \begin{tabular}{l|ccc|ccc|ccc}
        \toprule 
        & \multicolumn{3}{c|}{\textbf{Reddit}} & \multicolumn{3}{c|}{\textbf{Twitter}} & \multicolumn{3}{c}{\textbf{Zhihu}} \\
        & Object & Content & Type & Object & Content & Type & Object & Content & Type \\
        \midrule 
        \rowcolor{gray!20} \multicolumn{10}{c}{\textbf{Mistral-7b-Instruct}} \\
        Zero-Shot & 25.95±2.0 & 19.62±1.4 & 22.92±0.7 & 19.99±2.4 & 27.04±6.5 & 62.92±0.9 & 21.56±1.2 & 19.53±1.6 & 10.38±0.2 \\
        Few-Shot & 19.99±0.0 & 14.18±0.0 & 24.01±0.0 & 38.80±0.0 & 22.47±0.0 & 56.20±0.0 & 20.32±0.4 & \underline{23.51±1.2} & 14.51±0.2 \\
        std-CoT-FT & \underline{31.56±0.0} & 19.46±0.1 & \underline{31.40±0.0} & \underline{55.76±0.0} & \underline{55.56±0.0} & \underline{84.14±0.0} & \underline{29.30±0.4} & 16.61±0.8 & \underline{17.77±0.0} \\
        OM-CoT-FT(ours) & \textbf{34.58±0.2} & \textbf{21.27±0.2} & \textbf{45.12±0.7} & \textbf{64.38±0.0} & \textbf{56.25±0.0} & \textbf{88.12±0.0} & \textbf{33.19±0.7} & \textbf{28.66±0.5} & \textbf{21.54±0.1} \\
        \midrule 
        \rowcolor{gray!20} \multicolumn{10}{c}{\textbf{Solar-10.7b-Instruct}} \\
        Zero-Shot & 27.31±1.6 & \textbf{25.31±2.3} & 20.52±0.8 & 53.40±1.6 & 27.72±7.6 & 71.12±1.4 & 19.48±1.6 & 15.00±2.1 & 10.66±0.5 \\
        Few-Shot & 18.31±0.0 & 16.44±0.0 & 21.30±0.0 & 38.84±0.0 & 22.74±0.0 & 56.20±0.0 & 19.94±0.1 & 11.26±0.0 & 12.46±0.2 \\
        std-CoT-FT & \underline{28.30±0.0} & 21.20±0.1 & \underline{45.33±0.0} & \underline{62.27±0.0} & \textbf{58.86±0.0} & \underline{74.04±0.0} & \underline{24.63±0.3} & \underline{34.21±0.8} & \underline{16.18±0.4} \\
        OM-CoT-FT(ours) & \textbf{38.16±0.4} & \underline{23.23±0.0} & \textbf{47.44±0.1} & \textbf{73.53±0.0} & \underline{49.04±0.0} & \textbf{87.29±0.0} & \textbf{28.45±0.1} & \textbf{36.65±0.4} & \textbf{19.53±0.0} \\
        \bottomrule 
    \end{tabular}
    }
\end{table}

\subsection{Discussion}
In this subsection, we will conduct ablation studies to further analyze the key factors that influence behavior simulation performance. Specifically, we seek to address the following three research questions.

\begin{table}
    \caption{Ablation study on different prompt components to explore how various aspects influence the simulation of fine-grained behavior elements.}
    \label{table:roleinfo}
    \centering
    \begin{tabular}{l|l|l|l|l}
        \toprule
        & & \textbf{Object} & \textbf{Content} & \textbf{Type} \\
        \midrule
        \multirow{4}{*}{\makecell{Mistral-7b \\ (Zero-Shot)}} & ALL & 19.99 & 27.04 &  62.92 \\
        & w/o userinfo  & 17.59\redtext{$_{-2.4}$} & 17.63\redtext{$_{-9.4}$} &  53.81\redtext{$_{-9.1}$} \\
        & w/o interest   & 19.75\redtext{$_{-0.2}$}& 24.04\redtext{$_{-3.0}$} &  54.60\redtext{$_{-8.3}$} \\
        & w/o history & 11.99\redtext{$_{-6.0}$}& 21.14 \redtext{$_{-5.9}$} & 26.21\redtext{$_{-36.7}$} \\
        \midrule
        \multirow{4}{*}{\makecell{Mistrail-7b \\ (OM-Cot-FT)}} & ALL & 64.38 & 56.25& 88.12 \\
        & w/o userinfo &   64.61\greentext{$_{+0.3}$}&46.19\redtext{$_{-10.1}$}&86.60\greentext{$_{+0.5}$} \\
        & w/o interest &   65.49\greentext{$_{+0.1}$}&50.48\redtext{$_{-5.8}$}&86.06\redtext{$_{-2.1}$} \\
        & w/o history &   39.60\redtext{$_{-24.7}$}&43.12\redtext{$_{-13.17}$}&43.42\redtext{$_{-44.8}$} \\
        \bottomrule
    \end{tabular}
\end{table}

\paragraph{RQ1: Which part of the prompt is the most important?}
Behavior simulation prompt includes role's basic information, interests, and past behaviors.  To assess the importance of each component, we conduct ablation experiments by removing individual parts from the input prompts. Table \ref{table:roleinfo} shows the results on the Twitter dataset, demonstrating how these components affect model performance.
As highlighted in our preliminary experiments, role history is the most influential, especially for OM-CoT fine-tuned models that are trained to analyze historical behaviors. Removing role history leads to a notable performance drop. On the other hand, the effect of basic information and interests varies across different behavior elements. For example, excluding basic info and interests has minimal impact on simulating behavior object and type, but they are useful for accurately simulating behavior content. Notably, while OM-CoT emphasizes observation and memory-based reasoning, it still integrates character profile analysis within the CoT process, which relies on role information in the input prompt.

\begin{figure*}[htbp]
    \centering
    \includegraphics[width=\linewidth]{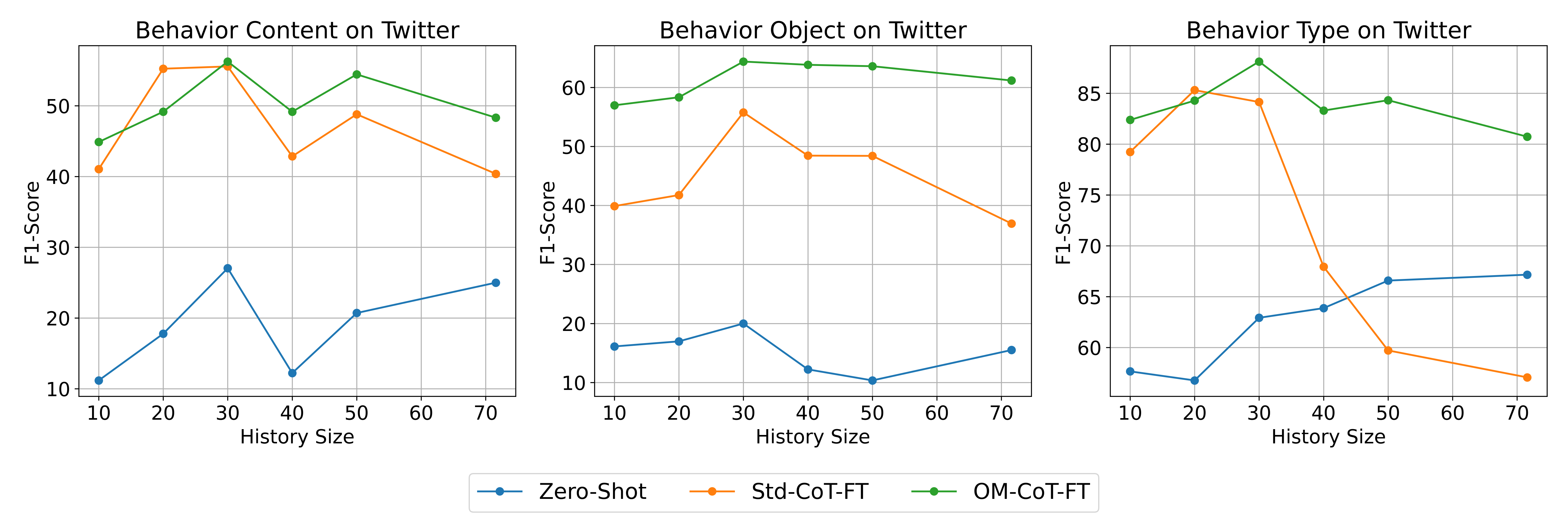}
    \caption{The relationship between input historical behavior size and the accuracy of simulating fine-grained behavior elements. The figure presents the results of three methods on the Twitter dataset.}
    \label{fig:history}
\end{figure*}

\paragraph{RQ2: Does adding more user history input improve the accuracy of behavior simulation?}
In the main experiment, we consistently choose the 30 most recent behavior history as the input. However, a plausible hypothesis suggests that including more behavior history could enhance behavior simulation, provided it fits within the model's token limit. To explore this, we evaluate the performance across different history window sizes, ranging from 10 to all entries(average 74)\footnote{We carefully clean each behavior’s text to avoid exceeding the token limitation of LLMs}, as shown in Figure \ref{fig:history}. Contrary to intuition, \textbf{adding more user behavior history does NOT consistently improve behavior simulation.}  We find that performance peaks at around 30 behavior entries, with additional history leading to a decline in accuracy. We hypothesize that increasing historical data introduces more noise, making it harder for the model to focus on relevant information. This finding aligns with human decision-making, where recent actions tend to be more influential. Interestingly, OM-CoT-FT models display greater stability with increased input history, showing promise for handling longer behavior sequences.


\begin{table}
    \caption{The ablation experiment of two special tokens used in our OM-COT-FT method. The table presents the results using the Mistral-7b-instruct model on the Twitter dataset.}
    \label{table:ablation}
\centering
\begin{tabular}{l|l|l|l}
\toprule
 & \textbf{Object} & \textbf{Type} & \textbf{Content} \\ \midrule
OM-CoT-FT & 64.38 & 56.25 & 88.12 \\
only \textbf{\texttt{<ANA>}} & 61.36\redtext{$_{-3.0}$} & 46.46\redtext{$_{-9.8}$} & 84.37\redtext{$_{-3.8}$} \\
only \textbf{\texttt{<MEM>}} & 58.88\redtext{$_{-5.5}$} & 55.58\redtext{$_{-0.7}$} & 72.25\redtext{$_{-15.9}$} \\ 
\bottomrule
\end{tabular}
\end{table}

\paragraph{RQ3:Do both the <ANA> and <MEM> special token work effectively?}
To investigate this, we conducted ablation experiments by selectively removing content enclosed by the special tokens \textbf{\texttt{<ANA>}} (analysis) and \textbf{\texttt{<MEM>}} (memory) during the reasoning process. This was done by either adjusting the system prompt or excluding these tokens during the decoding process. The results are shown in Table \ref{table:ablation}. The experimental findings reveal that removing either \textbf{\texttt{<ANA>}} or \textbf{\texttt{<MEM>}} leads to a decrease in behavior simulation performance, highlighting the importance of both tokens in the CoT reasoning process. However, the influence of these tokens varies across different sub-tasks. For behavior type simulation, the model relies more on analyzing available candidate options (\textbf{\texttt{<ANA>}}), whereas behavior content simulation depends more on recalling and reproducing historical behaviors (\textbf{\texttt{<MEM>}}). In the case of behavior object simulation, both observation and memory are equally important. This suggests that each sub-task requires a different balance between reasoning based on current observations and past behavior records to achieve optimal performance.


\subsection{Ethics}
Current LLMs have achieved highly realistic role-playing abilities, sometimes indistinguishable from real human. This development unlocks potential applications in areas like companionship, entertainment, and education, where human-like agents could lead to significant progress. However, these capabilities also pose risks in social networks, such as the spread of propaganda, misinformation, and the proliferation of malicious bots.This study aims to enhance LLM' ability to accurately simulate human roles while also providing insights into identifying behavioral patterns and thinking styles of such LLM-driven social bots.

\section{Conclusion}
In this work, we introduce FineRob, a new fine-grained behavior simulation dataset designed to explore how LLMs understand and simulate real human behaviors. We collect user behavior history from three social media platforms and break down each behavior record into three key elements: object, type, and content. Preliminary experiment on FineRob reveals two dominant reasoning patterns: role stereotype-based, observation and memory-based. The latter proves to be more effective in behavior simulation, leading us to propose OM-CoT method that explicitly integrates observation and memory into the reasoning process. We conduct comprehensive experiments on nine mainstream LLMs and analyze key influencing factors in depth. By comparing OM-CoT with baseline methods, we confirm the effectiveness of our approach.

\begin{table}[h]
\caption{An example of OM-COT Finetuned Reasoning COT}
\label{appendix: reasoning-case}
\centering
\begin{tcolorbox}[colback=white, colframe=black, title=\centering Reasoning Case created by OM-COT-Finetuned Mistral-7b-Instruct on Reddit dataset]
\textcolor{blue}{\section*{<ANA>}}
The options presented reflect varying perspectives on the accessibility and limitations of financial services, particularly regarding ATM withdrawals and online transactions. Option (C) specifically mentions the inability to withdraw money using a debit card, which aligns with the user's concern about the limitations of their financial account. Option (D) discusses the limitations set by financial institutions regarding ATM withdrawals and daily debt purchases, which is relevant to the user's inquiry about their financial situation.
\textcolor{blue}{\section*{</ANA>}}
\textcolor{red}{\section*{<MEM>}}
- The user commented on the high fees associated with withdrawing money from an ATM, indicating a concern about the costs of accessing their funds. \\
- The user also expressed frustration over the limitations of their financial account, particularly regarding the inability to withdraw money without incurring fees. \\
- The user has engaged in discussions about the challenges of managing finances in Australia, which may relate to the limitations mentioned in option (D) regarding financial institution policies.
\textcolor{red}{\section*{</MEM>}}
\textbf{Therefore, the answer is \textbf{(C)}.}
\end{tcolorbox}

\end{table}

\begin{table}[hp]
\caption{An example of User info inf FineRob Dataset}
\centering
\begin{tcolorbox}[colback=blue!5!white, colframe=blue!75!black, title=\centering A behavior content QA record of Twitter]

\section*{Role Info:}
\begin{itemize}
    \item \textbf{Username:} celebrities
    \item \textbf{Description:} Welcome to your 15 seconds of fame! Just a bit of fun :)
\end{itemize}
\section*{Interests:}
\begin{itemize}
    \item Swachhsurvekshan
    \item Ogwugfood
    \item Foodapp
\end{itemize}
\section*{Behavior History:}

\begin{itemize}
    \item \textbf{Post:} \\
    \textit{Action Time:} 2020-08-06 13:13:54 \\
    \textit{Content:} Election 2020 \#PresidentialDebates \#PresidentTrump \#Biden \#USA \\
    \textit{Link:} https://t.co/2SqNnemss9
    
    \item \textbf{Like:} \\
    \textit{Action Time:} 2020-08-14 09:59:57 \\
    \textit{Object:} He is without question a leader who pushes risky ideas forward. Via:@jongertner \\
    \textit{Link:} https://t.co/ilyXah4F8n
    
    \item \textbf{Retweet:} \\
    \textit{Action Time:} 2020-08-14 10:02:52 \\
    \textit{Object:} RT @savanteum: He is without question a leader who pushes risky ideas forward.
\end{itemize}

\end{tcolorbox}

\label{appendix: record-case}
\end{table}

%
%
\bibliographystyle{splncs04}
\bibliography{custom}

\begin{thebibliography}{10}
\providecommand{\url}[1]{\texttt{#1}}
\providecommand{\urlprefix}{URL }
\providecommand{\doi}[1]{https://doi.org/#1}

\bibitem{DBLP:journals/corr/abs-2402-10618}
Enhancing role-playing systems through aggressive queries: Evaluation and improvement. CoRR  \textbf{abs/2402.10618} (2024)

\bibitem{DBLP:conf/acl/AhnLLKYLK24}
Ahn, J., Lee, T., Lim, J., Kim, J., Yun, S., Lee, H., Kim, G.: Timechara: Evaluating point-in-time character hallucination of role-playing large language models. In: {ACL} (Findings). pp. 3291--3325. Association for Computational Linguistics (2024)

\bibitem{DBLP:conf/icml/PetersonB0GR19}
Bourgin, D.D., Peterson, J.C., Reichman, D., Russell, S.J., Griffiths, T.L.: Cognitive model priors for predicting human decisions. In: {ICML}. Proceedings of Machine Learning Research, vol.~97, pp. 5133--5141. {PMLR} (2019)

\bibitem{DBLP:journals/corr/abs-2406-11683}
Chen, J., Zhu, X., Yang, C., Shi, C., Xi, Y., Zhang, Y., Wang, J., Pu, J., Zhang, R., Yang, Y., Feng, T.: Hollmwood: Unleashing the creativity of large language models in screenwriting via role playing. CoRR  \textbf{abs/2406.11683} (2024)

\bibitem{DBLP:conf/emnlp/ChenWJ0LCWL23}
Chen, N., Wang, Y., Jiang, H., Cai, D., Li, Y., Chen, Z., Wang, L., Li, J.: Large language models meet harry potter: {A} dataset for aligning dialogue agents with characters. In: {EMNLP} (Findings). pp. 8506--8520. Association for Computational Linguistics (2023)

\bibitem{DBLP:conf/iclr/DinanRSFAW19}
Dinan, E., Roller, S., Shuster, K., Fan, A., Auli, M., Weston, J.: Wizard of wikipedia: Knowledge-powered conversational agents. In: {ICLR} (Poster). OpenReview.net (2019)

\bibitem{DBLP:journals/corr/abs-2407-21783}
Dubey, A., Jauhri, A., Pandey, A., Kadian, A., Al{-}Dahle, A., Letman, A., Mathur, A., Schelten, A., Yang, A., Fan, A., Goyal, A., Hartshorn, A., Yang, A., Mitra, A., Sravankumar, A., Korenev, A., Hinsvark, A., Rao, A., Zhang, A., Rodriguez, A., Gregerson, A., Spataru, A., Rozi{\`{e}}re, B., Biron, B., Tang, B., Chern, B., Caucheteux, C., Nayak, C., Bi, C., Marra, C., McConnell, C., Keller, C., Touret, C., Wu, C., Wong, C., Ferrer, C.C., Nikolaidis, C., Allonsius, D., Song, D., Pintz, D., Livshits, D., Esiobu, D., Choudhary, D., Mahajan, D., Garcia{-}Olano, D., Perino, D., Hupkes, D., Lakomkin, E., AlBadawy, E., Lobanova, E., Dinan, E., Smith, E.M., Radenovic, F., Zhang, F., Synnaeve, G., Lee, G., Anderson, G.L., Nail, G., Mialon, G., Pang, G., Cucurell, G., Nguyen, H., Korevaar, H., Xu, H., Touvron, H., Zarov, I., Ibarra, I.A., Kloumann, I.M., Misra, I., Evtimov, I., Copet, J., Lee, J., Geffert, J., Vranes, J., Park, J., Mahadeokar, J., Shah, J., van~der Linde, J., Billock, J., Hong, J., Lee, J., Fu, J.,
  Chi, J., Huang, J., Liu, J., Wang, J., Yu, J., Bitton, J., Spisak, J., Park, J., Rocca, J., Johnstun, J., Saxe, J., Jia, J., Alwala, K.V., Upasani, K., Plawiak, K., Li, K., Heafield, K., Stone, K., et~al.: The llama 3 herd of models. CoRR  \textbf{abs/2407.21783} (2024)

\bibitem{DBLP:journals/corr/abs-2405-13362}
Ebrat, D., Rueda, L.: Lusifer: Llm-based user simulated feedback environment for online recommender systems. CoRR  \textbf{abs/2405.13362} (2024)

\bibitem{DBLP:conf/acl/GaoLZFW23}
Gao, J., Lian, Y., Zhou, Z., Fu, Y., Wang, B.: Livechat: {A} large-scale personalized dialogue dataset automatically constructed from live streaming. In: {ACL} {(1)}. pp. 15387--15405. Association for Computational Linguistics (2023)

\bibitem{DBLP:journals/corr/abs-2308-05884}
Gosling, T., Dale, A., Zheng, Y.: {PIPPA:} {A} partially synthetic conversational dataset. CoRR  \textbf{abs/2308.05884} (2023)

\bibitem{DBLP:conf/iclr/HuSWALWWC22}
Hu, E.J., Shen, Y., Wallis, P., Allen{-}Zhu, Z., Li, Y., Wang, S., Wang, L., Chen, W.: Lora: Low-rank adaptation of large language models. In: {ICLR}. OpenReview.net (2022)

\bibitem{DBLP:journals/corr/abs-2310-06825}
Jiang, A.Q., Sablayrolles, A., Mensch, A., Bamford, C., Chaplot, D.S., de~Las~Casas, D., Bressand, F., Lengyel, G., Lample, G., Saulnier, L., Lavaud, L.R., Lachaux, M., Stock, P., Scao, T.L., Lavril, T., Wang, T., Lacroix, T., Sayed, W.E.: Mistral 7b. CoRR  \textbf{abs/2310.06825} (2023)

\bibitem{kaplan2010users}
Kaplan, A.M., Haenlein, M.: Users of the world, unite! the challenges and opportunities of social media. Business horizons  \textbf{53}(1),  59--68 (2010)

\bibitem{DBLP:journals/corr/abs-2312-15166}
Kim, D., Park, C., Kim, S., Lee, W., Song, W., Kim, Y., Kim, H., Kim, Y., Lee, H., Kim, J., Ahn, C., Yang, S., Lee, S., Park, H., Gim, G., Cha, M., Lee, H., Kim, S.: {SOLAR} 10.7b: Scaling large language models with simple yet effective depth up-scaling. CoRR  \textbf{abs/2312.15166} (2023)

\bibitem{DBLP:conf/acl/LiuCFMM23}
Liu, S., Cho, H., Freedman, M., Ma, X., May, J.: {RECAP:} retrieval-enhanced context-aware prefix encoder for personalized dialogue response generation. In: {ACL} {(1)}. pp. 8404--8419. Association for Computational Linguistics (2023)

\bibitem{DBLP:conf/acl/Lu0ZZ24}
Lu, K., Yu, B., Zhou, C., Zhou, J.: Large language models are superpositions of all characters: Attaining arbitrary role-play via self-alignment. In: {ACL} {(1)}. pp. 7828--7840. Association for Computational Linguistics (2024)

\bibitem{DBLP:conf/naacl/MostafazadehCHP16}
Mostafazadeh, N., Chambers, N., He, X., Parikh, D., Batra, D., Vanderwende, L., Kohli, P., Allen, J.F.: A corpus and cloze evaluation for deeper understanding of commonsense stories. In: {HLT-NAACL}. pp. 839--849. The Association for Computational Linguistics (2016)

\bibitem{DBLP:conf/uist/ParkOCMLB23}
Park, J.S., O'Brien, J.C., Cai, C.J., Morris, M.R., Liang, P., Bernstein, M.S.: Generative agents: Interactive simulacra of human behavior. In: {UIST}. pp. 2:1--2:22. {ACM} (2023)

\bibitem{DBLP:conf/acl/SalemiMBZ24}
Salemi, A., Mysore, S., Bendersky, M., Zamani, H.: Lamp: When large language models meet personalization. In: {ACL} {(1)}. pp. 7370--7392. Association for Computational Linguistics (2024)

\bibitem{schlosser2020self}
Schlosser, A.E.: Self-disclosure versus self-presentation on social media. Current opinion in psychology  \textbf{31}, ~1--6 (2020)

\bibitem{DBLP:conf/emnlp/ShaoLDQ23}
Shao, Y., Li, L., Dai, J., Qiu, X.: Character-llm: {A} trainable agent for role-playing. In: {EMNLP}. pp. 13153--13187. Association for Computational Linguistics (2023)

\bibitem{wang2023user}
Wang, L., Zhang, J., Yang, H., Chen, Z., Tang, J., Zhang, Z., Chen, X., Lin, Y., Song, R., Zhao, W.X., et~al.: User behavior simulation with large language model based agents. arXiv preprint arXiv:2306.02552  (2023)

\bibitem{DBLP:conf/coling/WangDGL24}
Wang, X., Dai, H., Gao, S., Li, P.: Characteristic {AI} agents via large language models. In: {LREC/COLING}. pp. 3016--3027. {ELRA} and {ICCL} (2024)

\bibitem{DBLP:conf/emnlp/WangCC23}
Wang, Z., Chiu, Y., Chiu, Y.C.: Humanoid agents: Platform for simulating human-like generative agents. In: {EMNLP} (Demos). pp. 167--176. Association for Computational Linguistics (2023)

\bibitem{DBLP:journals/corr/abs-2404-12138}
Xu, R., Wang, X., Chen, J., Yuan, S., Yuan, X., Liang, J., Chen, Z., Dong, X., Xiao, Y.: Character is destiny: Can large language models simulate persona-driven decisions in role-playing? CoRR  \textbf{abs/2404.12138} (2024)

\bibitem{DBLP:journals/corr/abs-2309-10305}
Yang, A., Xiao, B., Wang, B., Zhang, B., Bian, C., Yin, C., Lv, C., Pan, D., Wang, D., Yan, D., Yang, F., Deng, F., Wang, F., Liu, F., Ai, G., Dong, G., Zhao, H., Xu, H., Sun, H., Zhang, H., Liu, H., Ji, J., Xie, J., Dai, J., Fang, K., Su, L., Song, L., Liu, L., Ru, L., Ma, L., Wang, M., Liu, M., Lin, M., Nie, N., Guo, P., Sun, R., Zhang, T., Li, T., Li, T., Cheng, W., Chen, W., Zeng, X., Wang, X., Chen, X., Men, X., Yu, X., Pan, X., Shen, Y., Wang, Y., Li, Y., Jiang, Y., Gao, Y., Zhang, Y., Zhou, Z., Wu, Z.: Baichuan 2: Open large-scale language models. CoRR  \textbf{abs/2309.10305} (2023)

\bibitem{young2024yi}
Young, A., Chen, B., Li, C., Huang, C., Zhang, G., Zhang, G., Li, H., Zhu, J., Chen, J., Chang, J., et~al.: Yi: Open foundation models by 01. ai. arXiv preprint arXiv:2403.04652  (2024)

\bibitem{DBLP:journals/corr/abs-2402-13717}
Yu, X., Luo, T., Wei, Y., Lei, F., Huang, Y., Peng, H., Zhu, L.: Neeko: Leveraging dynamic lora for efficient multi-character role-playing agent. CoRR  \textbf{abs/2402.13717} (2024)

\bibitem{DBLP:conf/eacl/ZhaoZLZLHG24}
Zhao, R., Zhang, W., Li, J., Zhu, L., Li, Y., He, Y., Gui, L.: Narrativeplay: Interactive narrative understanding. In: {EACL} (Demonstrations). pp. 82--93. Association for Computational Linguistics (2024)

\bibitem{zheng2024llamafactory}
Zheng, Y., Zhang, R., Zhang, J., Ye, Y., Luo, Z., Feng, Z., Ma, Y.: Llamafactory: Unified efficient fine-tuning of 100+ language models. In: Proceedings of the 62nd Annual Meeting of the Association for Computational Linguistics (Volume 3: System Demonstrations). Association for Computational Linguistics, Bangkok, Thailand (2024), \url{http://arxiv.org/abs/2403.13372}

\bibitem{DBLP:journals/corr/abs-1901-09672}
Zheng, Y., Chen, G., Huang, M., Liu, S., Zhu, X.: Personalized dialogue generation with diversified traits. CoRR  \textbf{abs/1901.09672} (2019)

\bibitem{DBLP:journals/corr/abs-2311-16832}
Zhou, J., Chen, Z., Wan, D., Wen, B., Song, Y., Yu, J., Huang, Y., Peng, L., Yang, J., Xiao, X., Sabour, S., Zhang, X., Hou, W., Zhang, Y., Dong, Y., Tang, J., Huang, M.: Characterglm: Customizing chinese conversational {AI} characters with large language models. CoRR  \textbf{abs/2311.16832} (2023)

\end{thebibliography}
%





\end{document}